\documentclass[11pt,letterpaper]{article}
\usepackage{naaclhlt2013}
\usepackage{times}
\usepackage{latexsym}
\usepackage[T1]{fontenc}
\usepackage[cm]{sfmath}
\usepackage[utf8x]{inputenc}
\usepackage{float}
\usepackage{amsfonts,amstext,amsmath,amssymb}
\usepackage{graphicx,xspace}
\usepackage{subfig}
\usepackage{multirow}
\usepackage{array,url}
\usepackage{color}

\renewcommand{\Re}{\mathbb{R}}

\newcommand{\bbw}{\boldsymbol{w}}
\newcommand{\bbW}{\boldsymbol{W}}
\newcommand{\bbh}{\boldsymbol{h}}
\newcommand{\bbt}{\boldsymbol{t}}
\newcommand{\bbf}{\boldsymbol{f}}

\newcommand{\rel}{r}
\newcommand{\bbrel}{\boldsymbol{\rel}}

\newcommand{\card}[1]{|#1|}

\newcommand{\fb}{Freebase\xspace}

\newcommand{\wn}{Wordnet\xspace}

\title{Connecting Language and Knowledge Bases with Embedding Models\\ 
	for Relation Extraction }

\author{Jason Weston\\
	    Google\\
      111 8th avenue\\
      New York, NY, USA\\
	    \texttt{\small jweston@google.com}
	  \And
	Antoine Bordes\\
  	Heudiasyc UMR CNRS 7253\\
  	U. Tech. de Compi\`egne\\
  	Compi\`egne, France\\
  \texttt{\small bordesan@utc.fr}
  \And
  Oksana Yakhnenko\\
      Google\\
      111 8th avenue\\
      New York, NY, USA\\
      \texttt{\small oksana@google.com}
    \And
  Nicolas Usunier\\
    Heudiasyc UMR CNRS 7253\\
    U. Tech. de Compi\`egne\\
    Compi\`egne, France\\
  \texttt{\small usuniern@utc.fr}
  }

\date{}

\begin{document}
\maketitle
\begin{abstract}
\vspace{-1mm}
This paper proposes a novel approach for relation extraction from free text
which is trained to jointly use
information from the text {\em and} from existing knowledge.
Our model is based on two scoring functions that operate by learning low-dimensional embeddings of words and of 
entities and relationships from a knowledge base.
We empirically show on New York Times articles aligned with Freebase relations
 that our approach is able to efficiently use the extra information provided by a
large subset of Freebase data (4M entities, 23k relationships)
to improve over existing methods that rely on text features alone.

\end{abstract}

\vspace{-2mm}
\section{Introduction}
\vspace{-1mm}

Information extraction (IE) aims at generating structured data from free text
in order to populate Knowledge Bases (KBs).
Hence, one is given an incomplete KB
composed of a set of triples of the form $(h,\rel,t)$; $h$ is
the left-hand side entity (or {\em head}), $t$ the right-hand side entity (or {\em tail}) and $r$
the relationship linking them.
An example from the \fb KB\footnote{www.freebase.com} is (/m/2d3rf ,<director-of>, /m/3/324), 
where /m/2d3rf refers to the director ``Alfred Hitchcock" and /m/3/324 to the movie ``The Birds".

This paper focuses on the problem of learning to perform relation extraction (RE)
under weak supervision from a KB.
RE is sub-task of IE that considers that entities have already been detected by a different process,
such as a named-entity recognizer.
RE then aims at assigning to a relation mention $m$ (i.e. a sequence of text which states that some relation is true) 
the corresponding relationship from the KB, given a pair of extracted entities $(h,t)$ as context. 
For example, given the triplet (/m/2d3rf ,``wrote and directed", /m/3/324), a system should predict <director-of>.
%
The task is said to be weakly supervised because for each pair of entities $(h,t)$ detected in the text, 
all relation mentions $m$ associated with them are labeled with all the relationships connecting $h$ and $t$ in the KB,
whether they are actually expressed by $m$ or not.

Our key contribution is a novel
model that employs not only weakly labeled text mention data, as most
approaches do, but also leverages triples from the  known KB. 
The model thus 
learns the plausibility of new  $(h,\rel,t)$ triples by generalizing from the KB,
  even though this triple is not present. 
A ranking-based embedding framework is used to train such a model.
Thereby, relation mentions, entities and relationships are all embedded
into a common low-dimensional vector space, where scores are computed.
We show that our system can successfully take into account
 information from a large-scale KB (\fb: 4M entities,
23k relationships) 
to improve over existing systems, which are only using text features.

\vspace{-1mm}
\paragraph{Previous work}
Learning under weak supervision 
is common in Natural language processing,
especially for tasks
where the annotations costs are important
such as semantic parsing \cite{kate2007learning,liang2009learning,bordes2010label,Matuszek:12}.
This is also naturally used in IE, since it allows to train large-scale systems without requiring to label numerous texts.
The idea was introduced by \cite{craven1999constructing}, which matched the Yeast Protein Database with PubMed abstracts. It was also used to train open extractors based on Wikipedia infoboxes and corresponding sentences \cite{wu2007autonomously,wu2010open}.
%
%
Large-scale open IE projects \cite{banko2007open,carlson2010toward} also rely
 on weak supervision, since they learn models from a seed KB in order to extend it.

Weak supervision is also a popular option for RE: Mintz et
al.~\shortcite{mintz2009distant} used \fb to train weakly supervised
relational extractors on Wikipedia, an approach generalized by the
multi-instance learning frameworks
\cite{riedel2010modeling,hoffmann2011knowledge,surdeanu2012multi}. All
these works only use textual information to perform extraction.

%
%
Recently, Riedel et al.~\shortcite{riedel2013relation} proposed an approach to model jointly KB data and text by relying on collaborative filtering. Unlike our model, this method can not directly connect text mentions and KB relationships, but does it indirectly through joint learning of shared embeddings for entities in text and in the KB.
We did not compare with this recent approach, since it uses a different evaluation protocol than previous work in RE.


\section{Embedding-based Framework}
\vspace{-1mm}

Our work concerns energy-based methods, which learn low-dimensional vector representations ({\it embeddings}) of atomic symbols (words, entities, relationships, etc.). 
In this framework, we learn two models: one for predicting relationships given relation mentions and another one to encode the interactions among entities and relationships from the KB. 
The joint action of both models in prediction allows
us to use the connection between the KB and text to perform relation extraction.
One could also share parameters between models (via shared embeddings), but this is not implemented in this work.
%
%
This approach is inspired by previous work designed to connect words and \wn \cite{bordes:12aistats}.

Both submodels end up learning vector embeddings of symbols,
 either for  entities or relationships in the KB, or for each word/feature of the vocabulary 
(denoted ${\cal V}$). 
The set of entities and relationships in the KB are denoted by ${\cal
  E}$ and ${\cal R}$, and $n_v$, $n_e$ and $n_r$ denote the size of
${\cal V}$, ${\cal E}$ and ${\cal R}$ respectively. Given a triplet
$(h,r,t)$ the embeddings of the entities and the relationship (vectors in $\Re^k$) are denoted
with the same letter, in boldface characters (i.e. ${\bf h}$, ${\bf
  r}$, ${\bf t}$). 
\vspace{-1mm}
\subsection{Connecting text and relationships} \label{t2r}
\vspace{-1mm}

The first part of the framework concerns the learning of a function $S_{m2r}(m, \rel)$, based on embeddings, 
that is  designed
 to score the similarity of a relation mention $m$  and a relationship $r$. 
%

Our approach is inspired by previous work for connecting word labels and images \cite{weston2010large},
which we adapted, replacing images by mentions and word labels by relationships. 
Intuitively, it consists of first projecting windows of words into the embedding space and then computing a similarity measure (the dot product in this paper) between this projection and a relationship embedding.
The scoring function is then:
$$
S_{m2r}(m,\rel) = \bbf(m)^\top \bbrel
$$
with $\bbf$ a function mapping a window of words into $\Re^k$, $\bbf(m) = \bbW^\top\Phi(m)$; $\bbW$ is the matrix of $\Re^{n_v\times k}$ containing all word embeddings $\bbw$; $\Phi(m)$ is the (sparse)
binary representation of m ($\in \Re^{n_v}$)
%
%
and $\bbrel\in\Re^k$ is the embedding of the relationship $\rel$.


This approach can be easily applied at test time to score  (mention, relationship) pairs.
Since this learning problem is weakly supervised, Bordes et al. \shortcite{bordes2010label}
showed that a convenient way to train it is by using a ranking loss.
Hence, given a data set ${\cal D}= \{(m_i, \rel_i), i=1,\dots,\card{{\cal D}}\}$
 consisting of (mention, relationship) training pairs, 
one could  learn the embeddings using constraints of the form:
\begin{equation}\label{eq:cons1}
    \forall i,~ \forall \rel' \neq \rel_i, ~~ \bbf(m_i)^\top \bbrel_i  >    1 +  \bbf(m_i)^\top \bbrel'~~,
\end{equation}
where 1 is the margin.
Given any mention $m$ one can predict the corresponding relationship $\hat{r}(m)$ with:
\[
    \hat{r}(m)=\arg \max_{\rel'\in{\cal R}} S_{m2r}(m,\rel')=\arg \max_{\rel'\in{\cal R}}  \big(\bbf(m)^\top \bbrel' \big).
\]

Learning $S_{m2r}(\cdot)$ under constraints~\eqref{eq:cons1} is well suited when one is interested in building a 
a per-mention prediction system.
However, performance metrics of relation extraction are sometimes measured using
precision recall curves aggregated for all mentions concerning the same pair of entities, as in \cite{riedel2010modeling}.
In that case the scores across predictions for  different mentions need to be calibrated so that the most
confident ones have the higher scores. 
This can be better encoded 
with constraints of the following form:
\begin{equation*}
    \forall i,j,~~ \forall \rel' \neq \rel_i, ~~ \bbf(m_i)^\top \bbrel_i >  1 +  \bbf(m_j)^\top \bbrel'~~.
\end{equation*}
In this setup, scores of pairs observed in the training set should be larger than that of any other prediction across all mentions.
In practice, we use ``soft'' ranking constraints (optimizing the hinge loss), and enforce a (hard) constraint on the norms of the columns of $\bbW$ and $\bbrel$, i.e.
 $\forall_i, ||\bbW_i||_2 \leq 1$ and $\forall_j, ||\bbrel_j||_2 \leq 1$. 
Training  is carried out by stochastic gradient descent (SGD)
, updating $\bbW$ and $\bbrel$ at each step.
See \cite{weston2010large,bordes2013irreflexive} for details.

\vspace{-1mm}
\subsection{Encoding structured data of KBs} 
\vspace{-1mm}
Using only weakly labeled text mentions for training
ignores much of the prior knowledge we can leverage from a large KB such as Freebase.
In order to connect this
 relational data with our model,
we propose to  encode its information into entity and relationship embeddings.
This allows us to build a model which can score the plausibility of new entity relationship triples
which are missing from Freebase.
Several  models have been recently developed for that purpose (e.g. in \cite{Nickel:2011,bordesAAAI11,bordes:12aistats}): we chose in this work to use the approach of \cite{bordes2013irreflexive}, which is simple, flexible and has shown 
very promising results on \fb data.

Given a training set ${\cal S}=\{(h_i,\rel_i,t_i), i=1,\dots,\card{{\cal S}}\}$ of relations extracted from the KB, this model
learns vector embeddings of the entities and of the relationships using the 
idea that the functional relation induced by the
$\rel$-labeled arcs of the KB should correspond to a translation of the embeddings.
Hence, this method enforces that $\bbh+\bbrel \approx \bbt$ when $(h,\rel,t)$ holds,
while $\bbh+\bbrel$ should be far away from $\bbt$ otherwise. 
%
Hence such a model gives the following score
for the plausibility of a relation:
\begin{equation*}
S_{kb}(h,\rel,t) = - ||\bbh+\bbrel-\bbt||_2^2\,.
\end{equation*}
%

A ranking loss is also used
for training $S_{kb}$. 
The ranking objective is designed to assign higher scores
to existing relations versus any other possibility:
%
%
%
{{\small
\begin{eqnarray*}
    &\forall i,  \forall h' \neq h_i,&   S_{kb}(h_i,\rel_i,t_i) \geq  1 + S_{kb}(h',\rel_i,t_i),\\
    &\forall i,  \forall r' \neq r_i,&   S_{kb}(h_i,\rel_i,t_i) \geq  1 + S_{kb}(h_i,\rel',t_i),\\
    &\forall i,  \forall t' \neq t_i,&   S_{kb}(h_i,\rel_i,t_i) \geq  1 + S_{kb}(h_i,\rel_i,t').
\end{eqnarray*}
}}
As in section \ref{t2r} we use soft constraints,  enforce constraints on the norm of embeddings,
i.e. $\forall_{h,r,t}, ||h||_2 \leq 1, ||r||_2 \leq 1, ||t||_2 \leq 1$, and training is performed using SGD, as in 
%
\cite{bordes2013irreflexive}.

At test time, one may again need to calibrate the scores $S_{kb}$ across entity pairs.
We propose a simple approach:
we convert the scores by ranking all relationships ${\cal R}$ by $S_{kb}$ and instead
output:
\begin{equation*}
\tilde{S}_{kb}(h,r,t) = \Phi\big(\sum_{r' \neq r} \delta(S_{kb}(h,r,t) > S_{kb}(h,r',t))\big),
\end{equation*}
i.e. a function of the rank of $r$.  
We chose the simplified model $\Phi(x) = 1$ if $x < t$ and 0 otherwise.



\vspace{-1mm}
\subsection{Implementation for relation extraction}\label{sec:re}
\vspace{-1mm}

Our framework can be used for relation extraction in the following
way. First, for each pair of entities $(h,t)$ that appear in the test
set, all the corresponding mentions ${\cal M}_{h,t}$ in the test set
are collected and a prediction is performed with:
\begin{equation*}
\hat{\rel}_{h,t} = \mathop{\mathrm{argmax}}\limits_{\rel \in {\cal R}} \sum_{m\in{\cal M}_{h,t}} S_{m2r}(m,\rel)\,.
\end{equation*}
The predicted relationship can either be a valid
relationship or NA -- a marker that means that there is no relation
between $h$ and $t$ (NA is added to ${\cal R}$ during
 training and is treated like other relationships).
If $\hat{\rel}_{h,t}$ is a relationship, a composite
score is defined:
\begin{equation*}
  S_{m2r\text{+}kb}(h, \hat{\rel}_{h,t}, t)\! = \hspace{-0.3cm}\sum_{m\in{\cal M}_{h,t}}\hspace{-0.3cm}S_{m2r}(m,\hat{\rel}_{h,t}) + \tilde{S}_{kb}(h,\hat{\rel}_{h,t}, t) 
\end{equation*}
Hence, the composite model favors predictions that agree with both the
mentions and the KB. If $\hat{\rel}_{h,t}$ is NA, the score is unchanged.

%

\vspace{-1mm}
\section{Experiments}
\vspace{-1mm}

We use the training and test data, evaluation framework and baselines from
 \cite{riedel2010modeling,hoffmann2011knowledge,surdeanu2012multi}.

\vspace{-1mm}
\paragraph{NYT+FB}
This dataset, developed by \cite{riedel2010modeling}, aligns Freebase relations with the
New York Times 
corpus. Entities were found using the Stanford named entity tagger \cite{finkel2005incorporating}, 
and 
were matched to their name in Freebase.
For each mention, sentence level features are extracted which include part of speech,
named entity and dependency tree path properties. Unlike some of the previous methods, we do
not use 
features that 
aggregate properties across multiple mentions.
We kept the 100,000 most frequent features.
%
%
 There are 52 possible relationships and 121,034 training mentions of which 
 most are labeled as no relation (labeled ``NA'') -- there are 
 4700 Freebase relations mentioned in the training set, and 1950  in the test set.

\vspace{-1mm}
\paragraph{\fb}
\fb is a large-scale KB that has around 80M entities, 23k relationships and 1.2B relations.
We used a subset restricted to the top 4M entities (with the largest number of relations in a preprocessed subset) 
for scalability reasons.
We used all the 23k relationships.
To make a realistic setting, we did not choose the entity set using the NYT+FB data set, so it may not overlap completely.
For that reason, we needed to keep the set rather large.
Keeping the top 4M entities gives an overlap of 80\% with the entities in the NYT+FB test set.
Most importantly, we then removed all the entity pairs present in the  NYT+FB test set from \fb,
 i.e. all relations they are involved in independent of the relationship.
This ensures that we cannot just memorize the true relations for an entity pair --
 we have to learn to generalize from other entities and relations.

As the NYT+FB dataset was built on an earlier version of \fb we also had to translate the deprecated
 relationships into their new variants (e.g. 
``/p/business/company/place\_founded '' $\rightarrow$ ``/organization/organization/place\_founded'') 
to make the two datasets link (the 52 relationships in NYT+FB are now a subset of the 23k from \fb).
We then trained the $S_{kb}$ model on the remaining triples.

\vspace{-1mm}
\paragraph{Modeling}

Following \cite{bordes2013irreflexive} we set the embedding dimension $k$ to 50.
The learning rate for SGD was selected using a validation set: we obtained $0.001$ for $S_{m2r}$, and $0.1$ for $S_{kb}$.
For the calibration of $\hat{S}_{kb}$, $t=10$ (note, here we are ranking all 23k \fb relationships).
%
%
Training $S_{m2r}$ took 5 minutes, whilst training $S_{kb}$ took 2 days due
to the large scale of the data set.

%

\begin{figure}[t!]
\vspace{-5mm}
	\begin{center}
		\includegraphics[width=1\linewidth]{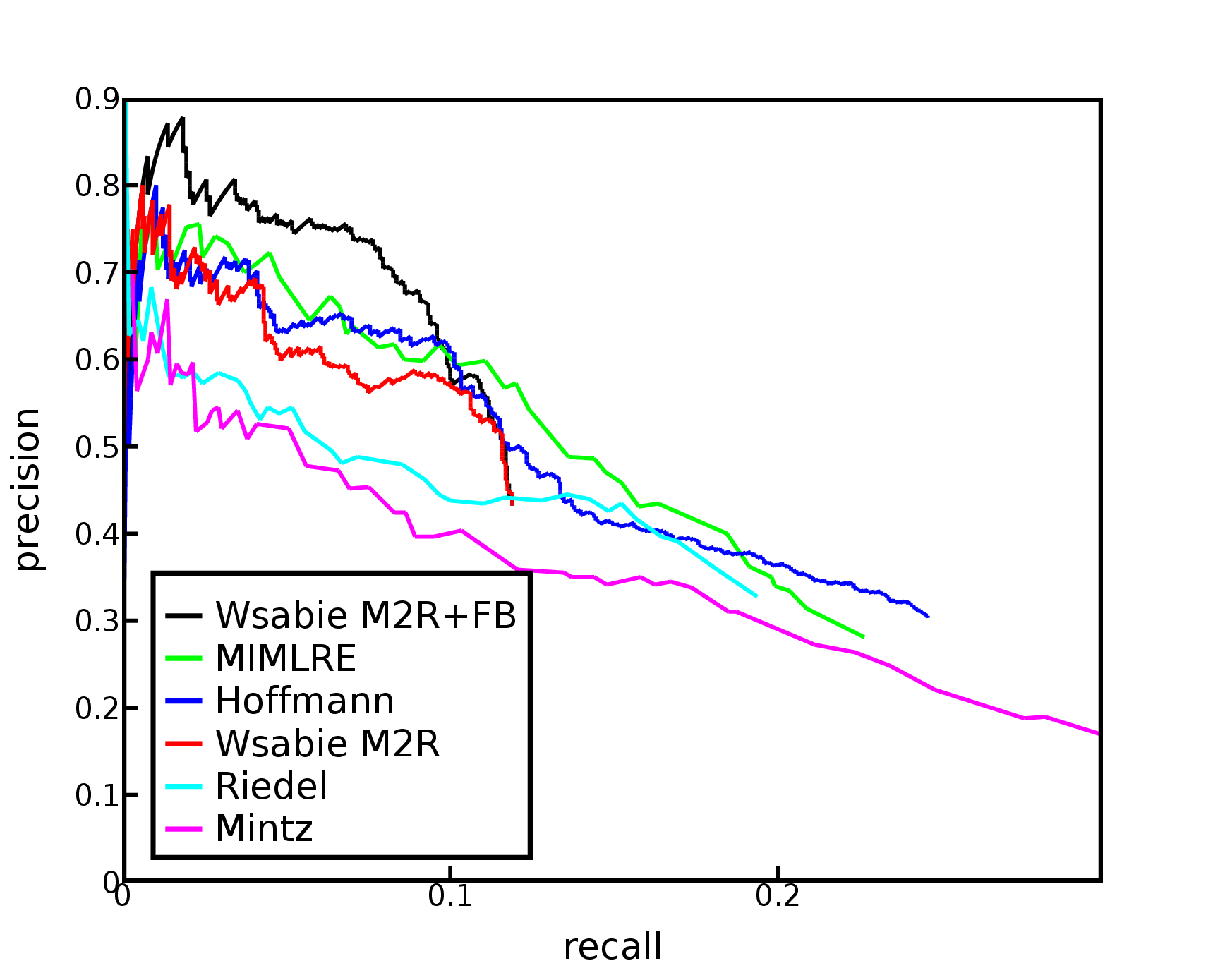}\vspace{-5mm}
		\includegraphics[width=1\linewidth]{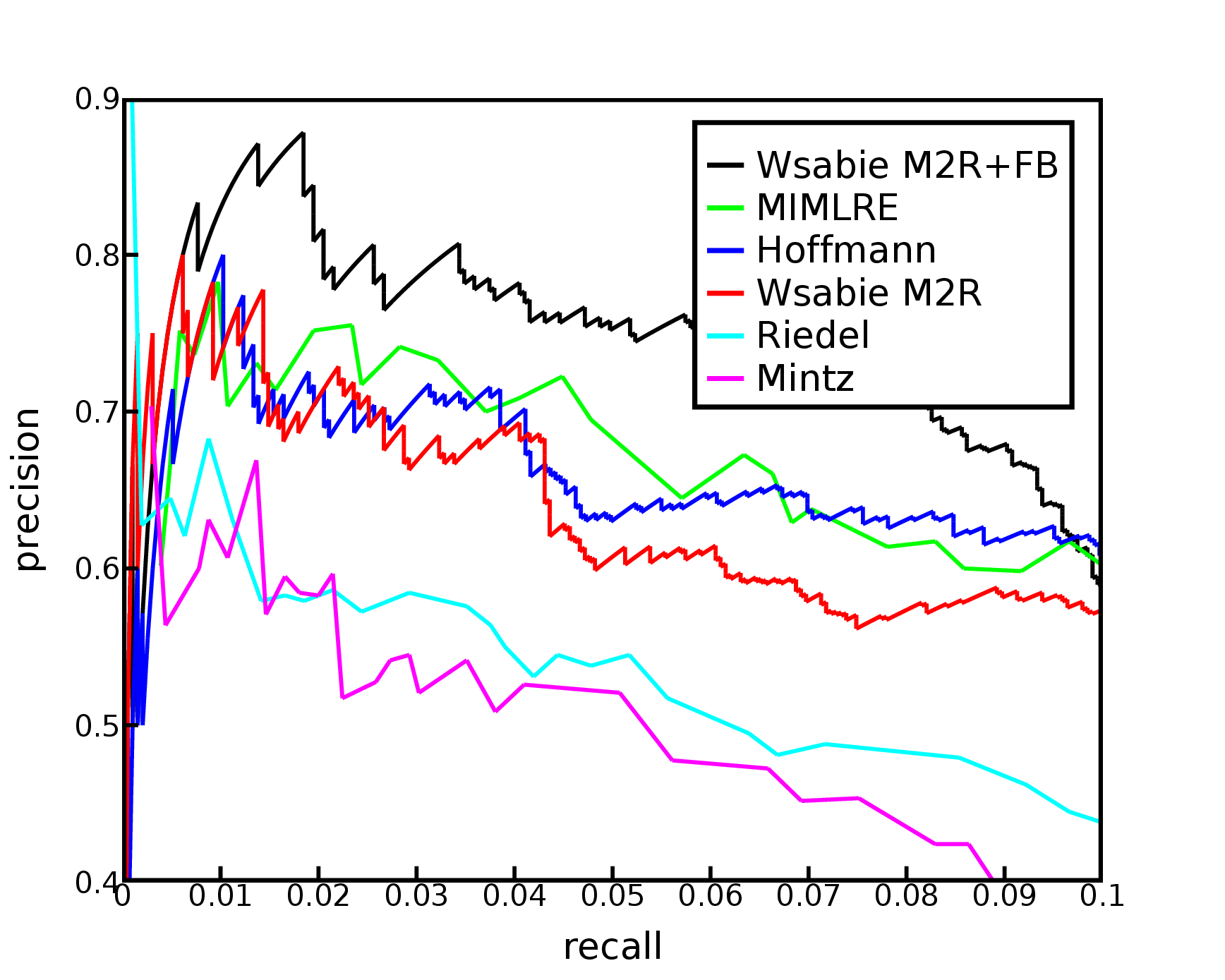}\vspace{-5mm}
	\caption{Top: Aggregate extraction precision/recall curves for 
a variety of methods.
 Bottom: the  same plot zoomed  to the recall [0-0.1] region.
Wsabie$_{M2R}$ is our method trained only on mentions, Wsabie$_{M2R+FB}$ uses Freebase
annotations as well. 
\label{fig1}
}
 \vspace{-5mm}
	\end{center}
\end{figure}

\vspace{-1mm}
\paragraph{Results}

Figure~\ref{fig1} displays the aggregate precision / recall curves of our approach 
{\sc Wsabie$_{M2R+FB}$} which uses the combination of $S_{m2r} + S_{kb}$, as well as
{\sc Wsabie$_{M2R}$ }, which only uses $S_{m2r}$, and state-of-the-art: 
 {\sc Hoffmann}~\cite{hoffmann2011knowledge}\footnote{There is an error
in the plot from \cite{hoffmann2011knowledge}, which we have corrected.
The authors acknowledged the issue.
}, {\sc mimlre}~\cite{surdeanu2012multi}.
{\sc Riedel}~\cite{riedel2010modeling} and
{\sc Mintz}~\cite{mintz2009distant}.

{\sc Wsabie$_{M2R}$ } is comparable to, but slightly worse than, the {\sc mimlre} and {\sc Hoffmann} methods,
possibly due to its simplified assumptions (e.g. predicting a single relationship per entity pair).
However, the addition of extra knowledge from other \fb entities in
{\sc Wsabie$_{M2R+FB}$} provides superior performance to all other methods, by a wide margin, at least 
between 0 and 0.1 recall (see bottom plot). 

\vspace{-2mm}
\section{Conclusion}
\vspace{-1mm}


In this paper we  described a framework for leveraging
large scale knowledge bases to improve relation extraction
by training not only on (mention, relationship) pairs but using all other KB
triples as well.
Our modeling approach is general and should apply to other settings,
e.g. for the task of entity linking.

\bibliographystyle{naaclhlt2013}
\bibliography{the_plan}

\end{document}